\pdfoutput=1

\documentclass[11pt]{article}
\usepackage[colorinlistoftodos]{todonotes}
\usepackage{graphicx}
\usepackage{EACL2023}
\usepackage{booktabs}
\usepackage{times}
\usepackage{multirow}
\usepackage{blkarray}
\usepackage{latexsym}
\usepackage{arydshln} 
\usepackage[T1]{fontenc}
\usepackage{microtype}

\usepackage{inconsolata}
\usepackage{amsmath}
\usepackage{amsfonts}
\usepackage{amssymb}

\usepackage{xcolor}
\usepackage{color-edits}
\addauthor[Alexis]{ap}{cyan}
\addauthor[Ali]{am}{teal}
\addauthor[Katharina]{kw}{blue}
\addauthor[Enora]{er}{violet}
\addauthor[Luke]{lg}{blue}

\title{TAMS: Translation-Assisted Morphological Segmentation}

\author{Enora Rice${ }^{1}$ \quad Ali Marashian${ }^{1}$ \quad  Luke Gessler${ }^{1}$  \quad
       Alexis Palmer${ }^{1}$ \\ {\bf Katharina von der Wense}${}^{1,2}$ \\
    ${ }^{1}$University of Colorado Boulder \quad ${ }^{2}$ Johannes Gutenberg University Mainz  \\ \texttt{enora.rice@colorado.edu}}

\begin{document}
\maketitle
\begin{abstract}
Canonical morphological segmentation is the process of analyzing words into the standard (\textit{aka} underlying) forms of their constituent morphemes.
This is a core task in endangered language documentation, and NLP systems have the potential to dramatically speed up this process. 
In typical language documentation settings, training data for canonical morpheme segmentation is scarce, making it difficult to train high quality models. 
However, translation data is often much more abundant, and, in this work, we present a method that attempts to leverage translation data in the canonical segmentation task. We propose a character-level sequence-to-sequence model that incorporates representations of translations obtained from pretrained high-resource monolingual language models as an additional signal. Our model outperforms the baseline in a super-low resource setting but yields mixed results on training splits with more data. Additionally, we find that we can achieve strong performance even without needing difficult-to-obtain word level alignments. While further work is needed to make translations useful in higher-resource settings, our model shows promise in severely resource-constrained settings.
\end{abstract}

\section{Introduction}
Morphological segmentation is the task of breaking words into morphemes, the smallest semantic units of a language.
Morphemes can merge and change during word formation, and the precise morphological composition of a word is often obfuscated in its surface form. Segmentation can thus take two forms: surface/linear segmentation and canonical segmentation, which divides a word into the ``canonical'' forms of its morphemes (cf. Figure 1). One important motivation for automated canonical segmentation is to expedite the process of linguistic analysis, including the creation of Interlinear Glossed Text (IGT). 
IGT is a form of morphological annotation that typically adheres to the Leipzig glossing format \cite{Lehmann_1982}, a linguistic representation wherein each line of the target text is broken up into a transcription line, a gloss line (morphological annotation), and a translation line.
IGT is a crucial resource in endangered language documentation work, but it is costly and time-consuming to generate.
The task of morphological segmentation is a key component in glossing, and automated canonical segmentation could aid in this process.
Prior work has shown that automated methods have the potential to assist language documentation and revitalization \cite{Palmer_Moon_Baldridge_2009,moeller-etal-2020-igt2p,moeller-hulden-2021-integrating,chaudhary2022autolex,ahumada-etal-2022-educational}.
\begin{figure}[t]
\centering
\includegraphics[scale=0.4]{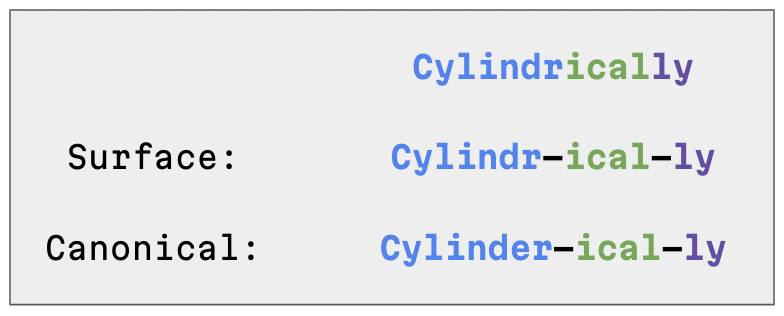}
\caption{Canonical segmentation of the English word "Cylindrically"}
\label{fig:cylinder}
\end{figure}

Neural models have been shown to perform well on the task of canonical segmentation \citep{Kann_Cotterell_Schütze_2016, ruzsics-samardzic-2017-neural}, but the success of these models has been restricted by the availability of annotated segmentation data.
IGT, though a limited resource, is one important source of training data for canonical segmentation.
Until now, primarily the transcription and gloss lines of IGT have been used as input to segmentation models, while the translations have been overlooked.
Moreover, in real-world language documentation settings, translated data is often much more available than morphologically analyzed data, making it practically attractive as an additional input. A typical language documentation pipeline begins with transcription, followed by translation, followed only then by morphological analysis. This commonly results in a setting where a linguist will have a wealth of translation data but comparatively little morphological segmentation data. Here we consider how supervised canonical segmentation methods can be improved by leveraging this underutilized supplemental data source.
Also, because they are typically into more widely-spoken languages, translations provide the opportunity to make use of pretrained models that likely have much higher-quality representations than any model available for the low-resource target language.

Our work is inspired in part by \citet{Zhao_Ozaki_Anastasopoulos_Neubig_Levin_2020} who experiment with leveraging translations for the task of automatic interlinear glossing (i.e., generating the gloss line of IGT). They use a multi-source word-level transformer to jointly model the transcription and translation sequences, and outperform previous baselines. Their work shows promise for the utility of translation data in morphological analysis tasks. However, they assume the presence of well-segmented data and state that "proper segmentation remains a challenge and that the creation of segmentation tools is a valuable endeavor." %\cite{Zhao_Ozaki_Anastasopoulos_Neubig_Levin_2020} 
Our work endeavors to address the segmentation issue.

We treat canonical segmentation as a sequence-to-sequence problem and use a character-level pointer-generator LSTM \cite{see-etal-2017-get} to map each surface form word to its segmented, canonicalized form.
We experiment with Tsez and Lezgi, two Northeast Caucasian languages, and Arapaho, a Plains Algonquian language. We leverage existing sentence-level English translations present in the IGT data from the SIGMORPHON 2023 Shared Task on Interlinear Glossing \cite{Ginn_Moeller_Palmer_Stacey_Nicolai_Hulden_Silfverberg_2023} and create two datasets of word-level transcription--translation alignments: automatically with awesome-align\footnote{Licensed under the BSD 3-Clause License.}\cite{Dou_Neubig_2021}, and manually according to conventions described in \S \ref{subsec:manual-align}.
We then embed these translations with 
BERT \cite{Devlin_Chang_Lee_Toutanova_2019} and experiment with incorporating them into our baseline model's encoder and decoder.  We also analyze the impact of training set size on the efficacy of our approach by limiting our training split to simulate varied levels of resourcedness. Finally, we analyze the effect of automatic vs. manual word-alignments on a subset of the Tsez data.
\footnote{\url{https://github.com/lgessler/tama/blob/master/data/tsez/wat/Tsez\%20Train\%20101-310.json}}
%the subset of the Tsez data that we manually aligned  
Based on poor automatic alignment performance, we introduce an additional model variant (\textbf{TAMS-CLS}) which incorporates the translation only at the sentence level.

We find that although gold alignment does lead to better performance with our approach, we still see improvement over the baseline with automatic alignments in some cases.  In the extremely low data setting (n=100), our approach outperforms the baseline by an average of 1.99 percentage points and as high as 2.87 points, even with poor quality automatic alignments. In many cases, across training set sizes, \textbf{TAMS-CLS} is the highest performing model configuration, which suggests that we may not even need word-level alignments to achieve performance improvements with \textbf{TAMS}. These results are promising in the context of the extremely resource constrained language documentation setting. However, in higher data settings, incorporating translations may or may not be beneficial. Our results also suggest that aligning translations may not be necessary to see improvements over the baseline.

\section{Related Work}
\paragraph{Modeling Morphological Segmentation}
Recent work on neural methods for canonical segmentation primarily focuses on LSTMs.
\citet{Kann_Cotterell_Schütze_2016} use a bidirectional RNN encoder-decoder with a neural reranker.
\citet{Mager_Çetinoğlu_Kann_2020} adapt this work to the low-resource setting and find that the pointer-generator network vastly improves over the performance of the LSTM canonical segmentation model for this setting. 
Recent work has also used the transformer for canonical segmentation: \citet{Moeng_Reay_Daniels_Buys_2021} test several sequence-to-sequence models for the task and find that the transformer performs the best on 4 Nguni languages.

\paragraph{Morphological Information within Embeddings}
Previous work has suggested that distributional 
similarity is an informative cue for morphology \citep{yarowsky-wicentowski-2000-minimally, schone-jurafsky-2001-knowledge}, and that static word embeddings encode some morphological information \citep{musil-2021-representations, soricut-och-2015-unsupervised}.
Other work has suggested that BERT embeddings could encode grammatical and morphological information \cite{nastase-merlo-2023-grammatical, jawahar-etal-2019-bert}.
BERT embeddings have also been used for part-of-speech tagging \cite{tsai-etal-2019-small, singh-etal-2021-pilot, Mohseni_Tebbifakhr_2019}. We aim to leverage the morphological cues intrinsic to pretrained embeddings of English translations to improve our segmentation models.

\section{Incorporating Translations into Morphological Segmentation Models}
\label{sec:approach}
Here, we present our method for using translations as a source of additional signal for morphological analysis.
We first align the word forms in our target language to relevant word forms in the translation. We then obtain embeddings of these word-forms from a high-resource language model.
We describe several approaches for turning these embeddings into a fixed-length representation and for incorporating them as input to the segmentation model.

\subsection{Encoder--Decoder Networks}
The most common architecture for morphological analysis is the neural encoder--decoder architecture with attention %\citep{sutskever_sequence_2014}.
\cite{bahdanau2015neural}.
An encoder--decoder network estimates the probability of an output sequence $\mathbf{y} = y_1, \ldots, y_{T^\prime}$ in terms of an input sequence $\mathbf{x} = x_1, \ldots, x_T$ by decomposing the output sequence's joint probability using the chain rule of probability, where $y_t$ is conditioned on all previous output items and some representation of the input sequence $v_t$ computed using a function $g$:
\[
\begin{aligned}
    p(y_1, \ldots, y_{T^\prime}) = &\ \prod_{t=1}^{T^\prime} p(y_t|v_t, y_1, \ldots, y_{t-1}) \\
    v_t = &\ \mathrm{g}(\mathbf{x}, y_1, \ldots, y_{t-1})
\end{aligned}
\]

For morphological tasks, this architecture is commonly implemented by treating words as character sequences and using RNNs for the encoder and the decoder.
The encoder is responsible for producing representations of $\mathbf{x}$ which are useful for the decoder, and the decoder is responsible for producing conditional probability distributions for making a prediction for $\mathbf{y}$, $\mathbf{\hat{y}}$.

\subsection{Translation Assistance}\label{subsec:trans}
The translations of textual data from low-resource languages are usually written in a high-resource language such as English or Spanish.
High-resource languages have very high-quality pretrained language models (PLMs) and therefore rich word representations available to them, and we hypothesize that incorporating this signal into the process of morphological segmentation may be helpful.
For example, information from the high-resource language might help the segmenter resolve lexical ambiguities.

Here, we propose several related methods for incorporating information from a high-resource translation into an RNN-based encoder--decoder morphological segmenter.
For clarity, we will concretely consider a unidirectional LSTM-based encoder, though our approach is trivially applicable to other RNN-based encoder--decoder architectures.
We refer to the network's embedding and hidden dimension sizes as $\mathrm{emb}$ and $\mathrm{hid}$, respectively.

Consider a translated sentence with $X$, $Y$, and $R$.
$X = \mathbf{x}_1, \ldots, \mathbf{x}_n$ is a sequence of unsegmented words.
$Y = \mathbf{y}_1, \ldots, \mathbf{y}_n$ is the sequence of the corresponding segmented words.
$R = \mathbf{r}_1, \ldots, \mathbf{r}_m$ is the sequence of words in the translation.
We use a PLM to obtain a dense representation for each translation word, $\mathbf{d} = d_1, \ldots, d_m$.
We additionally refer to any sentence-wide representation (such as BERT's \texttt{\small [CLS]} token) that the PLM might produce as $d_0$.
We refer to the PLM's hidden representation size as $\mathrm{h}_\mathrm{PLM}$.

\paragraph{Alignment}
A preliminary step is to produce alignments between source and translation words like so, where $\mathrm{align}$ represents an aligner's decision on whether two words are aligned:
\[ 
A = \{\langle \mathbf{x}_i, \mathbf{r}_j \rangle | \mathbf{x}_i \in X \wedge \mathbf{r}_j \in R \wedge \mathrm{align}(\mathbf{x}_i, \mathbf{r}_j) \} 
\]
The aligner is assumed to be external to the segmentation system.

\paragraph{Translation Representation}
For each word $\mathbf{x}$, we now have some aligned translation word representations $\mathbf{d}_\mathrm{align} =d_a, \ldots, d_b$.
We next produce $v$, a fixed-length representation of $\mathbf{d}_\mathrm{align}$ which will be of length $\mathrm{emb}$.
We investigate three different strategies for producing this representation which differ in how they treat the sentence-wide
representation $d_0$. The intuition behind including the CLS token in our representation is that it may allow us to capture sentence-level dynamics better than word-level alignments alone.
For the \textbf{CLS-None} strategy, we discard $d_0$ and average pool $\mathbf{d}_\mathrm{align}$ before using a model parameter $W_\mathrm{trans} \in \mathbb{R}^{\mathrm{h}_\mathrm{PLM} \times \mathrm{emb}}$ to project the vector from the hidden size of the PLM to the embedding size of the model:
\[
  v = \mathrm{avg}(d_a, \ldots, d_b) W_\mathrm{trans}
\]
The \textbf{CLS-Avg} strategy is identical to CLS-None except that $d_0$ is included in the average:
\[
  v = \mathrm{avg}(d_0, d_a, \ldots, d_b) W_\mathrm{trans}
\]
For the \textbf{CLS-Concat} strategy, we first average the aligned words like in CLS-None, but we introduce two model parameters, $W_\mathrm{trans}, W_\mathrm{cls} \in \mathbb{R}^{h_\mathrm{PLM} \times \frac{1}{2}\mathrm{emb}}$, where $W_\mathrm{trans}$ is applied to the averaged words and $W_\mathrm{cls}$ is applied to $\mathbf{d}_0$, and their concatenation is used as the final fixed vector:
\[
\begin{aligned}
  v_1 &= \mathrm{avg}(d_a, \ldots, d_b) W_\mathrm{trans} \\
  v_2 &= d_0 W_\mathrm{cls} \\ 
  v   &= v_1 \oplus v_2
\end{aligned}
\]
\paragraph{Incorporation Strategies}
After we have computed $v$, we need to incorporate it into the encoder and/or the decoder's process. We consider four different strategies for incorporation, described next.

For \textbf{Concat}, we double the model's input size to $2\times\mathrm{emb}$ and concatenate $v$ to the input embedding at each time step in the LSTM. 
\textbf{Concat-Half} is identical to Concat, except the model's input size is held constant, with character embeddings and $v$ sharing the dimensions equally.
The model's character embedding module and $W_\mathrm{trans}$ above have their output dimensions halved accordingly.
For \textbf{Init-State}, assuming that there is some integer $z$ such that $z \times \mathrm{emb} = \mathrm{hid}$, we initialize the LSTM's hidden state as $z$ concatenations of $v$ to itself, $\bigoplus_1^z v$.
For \textbf{Init-Char}, we modify the LSTM's input sequence so that $v$ appears first, as if it were the embedding of a character.

All strategies are applicable to either the encoder or the decoder; we experiment with most combinations, except for Init-Char in the decoder. 

\section{Data} 
To perform our experiments, we use IGT data from the SIGMORPHON 2023 Shared Task on Interlinear Glossing \cite{Ginn_Moeller_Palmer_Stacey_Nicolai_Hulden_Silfverberg_2023} in three languages: Lezgi, Tsez, and Arapaho. All of the data is licensed under CC BY-NC 4.0. 
Each word in this IGT format has a surface form, a canonical segmentation, morpheme-level glosses, and an English translation for the sentence the word appears in.
An example is shown in Figure \ref{fig:tsez igt}, which also provides a sample of our manual alignment process.

\subsection{Languages}

We experiment with three languages: Arapaho, Lezgi, and Tsez. 
All three languages present interesting modeling challenges given the complexity of their morphological processes.
In addition, the experiments cover different types of difficult morphology, as the two language families represented are typologically distinct.

\subsubsection{Tsez}
Tsez [ddo] belongs to the Tsezic subgroup, which is part of the larger Nakh-Daghestanian language family. Its morphology is highly agglutinative and suffixing. Tsez has complex nominal case morphology that allows multiple case suffixes to modify a single word, and there are around 250 possible combinations of these case suffixes. In terms of verbal morphology, Tsez separates verbs into four groups depending on the final segment of the stem, which affects the surface representation of the composite morphemes, including five possible indicative tense-aspect suffixes \cite{comrie_and_polinksy}. Tsez also has a rich set of converbs that are derived from the verb stem through multi-step morphophonological processes. We consider Tsez to be our development language and conduct all hyperparameter tuning on Tsez.

\subsubsection{Lezgi}
Lezgi [lez] belongs to the Lezgic branch within the Nakho-Daghestanian language family. Like, Tsez, Lezgi is a highly agglutinative language with a largely suffixing morphology. Lezgi morphology is predominantly inflectional and nouns are inflected for number, case, and localization. There are 18 nominal cases in Lezgi, 14 of which are locative \cite{Haspelmath_1993}. Morphologically, verb stems are divided into three groups -- Masdar, Imperfective, and Aorist stems -- which impact the inflectional suffixes they can take on. Three distinct verb forms can be derived from the Masdar stem, nine from the Imperfective, and five from the Aorist \cite{Haspelmath_1993}. Several additional secondary verbal categories particularly relating to mood can be achieved via suffixing on the verb. Given its close phylogenetic relationship to Tsez, we consider Lezgi to be an in-family test language.

\subsubsection{Arapaho}
Arapaho [arp] is an Algonquian language that is highly agglutinating and polysynthetic. Noun stems can be inflected for plurality, obviation, vocative, and locative cases through suffixing. Nouns also necessarily belong to either animate or inanimate gender, and gender impacts the surface representation of many inflectional markers. Arapaho nouns also participate in derivational morphology, and modified nouns can be derived from independent nouns or verbs. 

Arapaho verbal morphology is even more complex. In terms of inflectional morphology, verb stems can be divided into four different classes that each take different markers for person, number, and obviation. Verbs can also be broken up into four different orders-- affirmative, non-affirmative, conjunct, and imperative-- which also impact the inflectional morphemes. Arapaho derivational verbal morphology is extensive, and unique verb forms can be derived through processes of prefixation, suffixation, denominalization, reduplication, and noun incorporation. We consider Arapaho to be particularly interesting as 
an out-of-family test language
because of its rich morphology that is notably distinct from that of Tsez.

\subsection{Preprocessing}
The transcription and translation lines are not always pretokenized in these datasets (e.g., punctuation sometimes appears next to words), and it is necessary to tokenize the data for this reason.
We use HuggingFace transformers' \cite{Wolf_Debut_Sanh_Chaumond_Delangue_Moi_Cistac_Rault_Louf_Funtowicz_et_al._2020} \texttt{\small BertPreTokenize} pretokenizer for this purpose, with some additional processing to make language-specific corrections.
(In Arapaho, for example, the apostrophe character \textquotesingle{} represents a consonant, and it should not be separated from words it appears in.)
After processing, we verify that there are equal numbers of surface and canonical forms, and we discard any sentences for which this is not true.
Finally, we initialize our training instances by finding all unique pairs of surface and canonical forms at the word level and choose one randomly if there is more than one occurrence of it in the corpus.
Both surface and canonical forms are NFD normalized.
Our full datasets consist of 53800 words in Arapaho, 10952 words in Tsez, and 2060 words in Lezgi.
In our experiments, the Arapaho dataset is downsampled to 16666 words, to bring its size closer to the other two datasets.

\begin{figure}[t]
\centering
\includegraphics[scale=0.65]{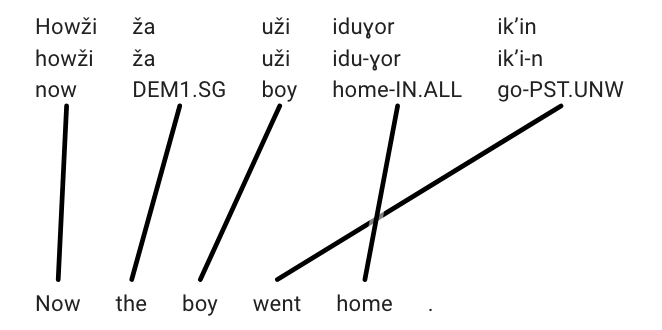}
\caption{Manual Word Alignment of Tsez IGT: \textit{Now the boy went home}}
\label{fig:tsez igt}
\end{figure}
\subsection{Word Alignment}\label{sec:align}
To facilitate canonical segmentation on the word level, we preprocess our dataset by aligning words in the transcription line to their corresponding word(s) in the translation line.
We experiment with two alignment methods: automatic and manual.

\subsubsection{Automatic Alignment}
We automatically align with awesome-align \cite{Dou_Neubig_2021}, which extracts word alignments from multilingual BERT and does not require training data for application to a new target language.
We use awesome-align's default hyperparameters except for the following: we use softmax extraction, set the softmax threshold to 5e-6, and set batch size to 32. 
After alignment, we split the instances for our main experiments into train, development, and test sets at a 60/20/20 ratio.

\subsubsection{Manual Alignment}
\label{subsec:manual-align}
We manually create gold alignments for the Tsez data according to the general principles laid out in \citet{Melamed_1998}. \citet{Melamed_1998} provides conventions for navigating complications that arise when translation is not literal, such as omissions, phrasal correspondence and idioms, amongst other linguistic nuances.
Additionally, we define several language-specific principles outlined in Appendix \ref{sec:appendixa} that address unique difficulties in mapping Tsez grammatical constructions to English. Figure \ref{fig:tsez igt} shows an example sentence in Tsez that we have manually aligned.
Our full gold-aligned dataset consists of 1419 words, which we divide into 500-word training and test sets, and a 419-word development set.

\section{Experiment}
We experiment with the strategies described in \S \ref{subsec:trans} for incorporating information from translations into our morphological segmentation model.
We proceed by first tuning our exact approach on the development split of a single language, Tsez, by exhaustively considering every combination of encoder, decoder, and CLS token translation incorporation strategies.
We then apply our top-performing model to the test splits of all three languages.
All experiments are performed on NVIDIA A100 GPUs, and all model implementations were based on those provided by Yoyodyne.\footnote{\url{https://github.com/CUNY-CL/yoyodyne}}

\subsection{Translation Vectors}
We use \texttt{bert-base-cased} \cite{Devlin_Chang_Lee_Toutanova_2019} to generate contextual word embeddings of the translations of each word in our aligned dataset. 
We then generate fixed-length translation vectors by averaging the embeddings of each word-piece in the translation sequence that was aligned to the word under consideration, as described in \S \ref{sec:approach}.

\subsection{Evaluation Metrics}
We employ three common metrics for evaluation. The first is whole-word accuracy, indicating the proportion of words that were segmented entirely correctly. To get a better picture of subword-level errors, we also use character-level edit distance. Finally, we use the modified F1 score outlined in \citet{Mager_Çetinoğlu_Kann_2020} to calculate the  F1 score at the morpheme level. We consider precision to be the proportion of morphemes in the prediction that also occur in the gold, and recall to be the proportion of morphemes in gold that also occur in the prediction.
\begin{table*}[t]
    \centering
    \scriptsize
    \begin{tabular}{cc|ccc|ccc|ccc|ccc} 
    & & \multicolumn{3}{c}{\textsc{Tsez}} & \multicolumn{3}{c}{\textsc{Lezgi}} & \multicolumn{3}{c}{\textsc{Arapaho}} & \multicolumn{3}{c}{\textsc{Average}}  \\ \midrule
    \textbf{Train} & \textbf{Metrics} & \textbf{TAMS} & \textbf{TAMS-} & \textbf{Base} & \textbf{TAMS} & \textbf{TAMS-} & \textbf{Base} & \textbf{TAMS} & \textbf{TAMS-} & \textbf{Base} & \textbf{TAMS} & \textbf{TAMS-} & \textbf{Base} \\
    \textbf{Limit} & & & \textbf{CLS} & & & \textbf{CLS} & & & \textbf{CLS} & & & \textbf{CLS} \\
    \midrule
    & Acc. & \textbf{24.78} & 23.83 & 22.19 & \textbf{29.71} & \textbf{29.71} & 26.84 & 15.05 &  \textbf{15.18} & 14.00 & \textbf{23.18} & 22.91 & 21.01 \\ 
    n = & F1 & \textbf{48.37} & 48.07 & 47.52 & 40.62 & \textbf{41.23} & 38.33 & \textbf{37.48} &  37.00 & 36.25 & \textbf{42.16} & 42.10 & 40.70 \\ 
    100 & ED & \textbf{3642} & 3701 & 3805 & 793 & \textbf{773} & 861 & 30759 & 30906  & \textbf{32742} & \textbf{11731} & 11793 & 12469 \\ \hline
    & Acc. & \textbf{34.52} & 34.01 & 33.36 & 34.80 & \textbf{35.63} & 35.19 & 20.66 &  \textbf{21.58} & 21.50 & 29.99 & \textbf{30.41} & 30.02 \\ 
    n = & F1 & 58.13 & \textbf{58.24} & 57.83 & 50.24 & \textbf{51.54} & 51.36 & 44.69 &  45.35 & \textbf{45.85} & 51.02 & 51.04  & \textbf{51.68} \\ 
    250 & ED & 3061 & \textbf{3055} & 3286 & 772 & \textbf{706} & 716 & 28561 & \textbf{27550}  & 29048 & 10798 & \textbf{10437} & 11017 \\ \hline
    & Acc. & 47.67 & 47.69 & \textbf{47.91} & 41.41 & 41.26 & \textbf{41.60} & 33.84 & \textbf{34.32} & 33.70 & 40.97 & \textbf{41.09} & 41.07 \\ 
    n = & F1 & 68.77 & 68.65 & \textbf{68.80} & \textbf{57.64} & 56.64 & 57.28 & 56.48 & \textbf{56.97} & 56.54 & \textbf{60.96} & 60.75 & 60.87 \\ 
    500 & ED & 2213 & \textbf{2195} & 2217 & \textbf{617} & 633 & 647 & \textbf{19743} & 19790 & 20211 & \textbf{7524} & 7539 & 7692 \\ \hline
    & Acc. & 80.78 &  81.96 & \textbf{82.60} & 46.84 & \textbf{47.09} & 44.66 & \textbf{67.72} & 67.40 & 67.08 &  65.11 & \textbf{65.48} & 64.78\\ 
    n = & F1 & 89.52 & 90.08  & \textbf{90.44} & \textbf{62.48} & \textbf{62.48} & 60.75 & \textbf{81.62} & 81.45 & 81.11 & 77.87 &  
    \textbf{78.00} & 77.43\\ 
    \textit{all} & ED & 701 & \textbf{643}  & 652 & \textbf{532} & 537  & 568 & \textbf{9899} & 9970 & 10495 & \textbf{3711} & 3717 & 3905\\
    \bottomrule
    \end{tabular}
    \caption{\textbf{Final results per language:} Performance on all languages' test sets (averaged over 5 randomized test sets) on silver-aligned data. Metrics: Accuracy (Acc.), F1 score (F1), and Edit Distance (ED).}
    \label{tab:combined-test}
\end{table*}

\subsection{Model and Hyperparameters}
\label{subsec:hyperparam}
In preliminary experiments, we conduct a hyperparameter search in order to determine which model architectures, model sizes, and optimization methods are most effective for these datasets.

\paragraph{Baselines and Settings for Hyperparameter Tuning}
Our baseline models for this task perform canonical segmentation without taking translations into account. The architectures we consider are a Transformer, a pointer-generator LSTM with a bidirectional encoder, and an attentive LSTM. Details are outlined in Appendix \ref{sec:appendixb}.
As it would be prohibitively expensive to do this in every experimental setting, we limit the scope of this search to baseline models  on the Tsez datasets on three training split sizes: 100, 500, and 6572.

\paragraph{Architectures} The pointer-generator LSTM \cite{see-etal-2017-get} performs better in all settings than either of the two other model architectures, and we therefore adopt it for all subsequent experiments.
Moreover, this choice is supported by evidence from \citet{Mager_Çetinoğlu_Kann_2020} that the pointer-generator is well suited to the low-resource canonical segmentation task. 

The pointer-generator LSTM differs from a regular LSTM encoder-decoder in that it has a pointer network \cite{Vinyals_Fortunato_Jaitly_2015}, which allows the model to copy over specific characters in the input sequence to the output sequence.
The decoder assesses the probability of copying an element from the input to the output rather than generating it, then computes the probability distribution of the output at each time step by combining the probability distribution across the output vocabulary with the attention distribution over the input characters. 
The weights, indicating the probability of generation or copying, are determined by a feedforward network. 

\paragraph{Data Size Matters}
The results of hyperparameter tuning are very similar for the 500-word and full data settings, but vary notably for the 100-sample setting.
In all subsequent experiments, we train our 100 training sample models with one set of hyperparameters and all other models with a separate set. 

In the 100-sample setting, we use a batch size of 16, two encoder and two decoder layers, an embedding size of 512 and a hidden size of 1024. 
We set dropout to $3.662 \times 10^{-1}$, learning rate to $2.411 \times 10^{-4}$, and train for up to 607 epochs.

For all other experimental settings, we use a batch size of 64, one encoder and one decoder layer, an embedding size of 1024, and a hidden size of 2048. 
We set dropout to $2.212 \times 10^{-1}$, learning rate to $8.056 \times 10^{-4}$, and train for up to 627 epochs. 

Both settings use the Adam optimizer and the ReduceLROnPlateau scheduler.\footnote{\url{https://pytorch.org/docs/stable/generated/torch.optim.lr\_scheduler.ReduceLROnPlateau.html.}}

\subsubsection{Alignment Considerations}
\label{sec:cls-only}

\begin{table}
    \centering
    \begin{tabular}{c|c|c}
         Precision&  Recall&   F1\\
         \hline
         0.1637&  0.1846&  0.1735\\
    \end{tabular}
    \caption{Performance of awesome-align judged against 500 manually aligned Tsez words}
    \label{tab:awesome-align quality}
\end{table}

Unfortunately, awesome-align does not produce alignments comparable to our gold-alignments. As reported in Table \ref{tab:awesome-align quality}, awesome-align achieves an F1 score of only 0.1735 on our set of 500 gold aligned Tsez samples. Though we cannot assess the alignment quality of the automatic alignments we produced for Lezgi or Arapaho, we take this as an indication that our automatic alignments are of dubious accuracy, and this may  adversely affect the performance of our approach. To address this possibility, we experiment with removing alignment from our approach entirely. We treat the cls-token embedding as a sentence-level representation of our translation input and use this in place of word-level embeddings. We then incorporate the embeddings as normal. We call this configuration \textbf{CLS-Only}.

\section{Results and Discussion}

We treat Tsez as our development language and perform a search over all translation incorporation strategies using the Tsez development set to determine the overall highest performing configuration, which we call \textbf{TAMS}. This configuration consists of \textbf{Init-State} in the encoder, \textbf{Concat-Half} in the decoder, and \textbf{CLS-Concat} as the CLS strategy.  We additionally consider a variation we call \textbf{TAMS-CLS}, which is identical to \textbf{TAMS} except that it employs the \textbf{CLS-Only} strategy described in \S  \ref{sec:cls-only}.  Further details of our translation incorporation strategy search are included in \S \ref{sec:translation incorporation}.

\subsection{Test Languages}
\label{subsec:test}
We apply our final \textbf{TAMS} model configuration to the automatically aligned test sets in each of our three languages: Tsez (development language), Lezgi (in-family test language), and Arapaho (out-of-family test language). To simulate varying levels of data availability, we experiment with models trained on 100, 250, and 500 training samples in addition to experimenting with the full training set. For each of the train-limits, we report average metrics over five randomly chosen subsets of the full training set. There was a relatively high level of variance in the performance of our models, which should be considered when interpreting our results. The  standard deviation of accuracy measures ranges between 1.11-2.33 across all settings. Full standard deviation metrics are reported in Appendix \ref{appendixc}.

Results are shown in Table \ref{tab:combined-test}. In most cases, we see that edit distance, F1, and accuracy are roughly in agreement, so we focus on accuracy as our main evaluation metric. We find that on average our \textbf{TAMS} and/or \textbf{TAMS-CLS} outperforms the baseline in every train data setting and almost every metric. In general, performance gains are highest on the lower-resource settings, while on the higher-resource settings, performance improvements are slight if present at all. 
In the n=100 setting, \textbf{TAMS} outperforms the baseline by an average of 1.99 percentage points, suggesting that our model is most beneficial in truly low-data settings. 
We also see consistent gains on Arapaho, which indicates that with further work, our model could be useful for polysynthetic languages. This is particularly exciting considering the relative difficulty of segmenting polysynthetic languages.

Without assessments of alignment quality for each of our test sets, we cannot properly analyze the impact of poor alignments on \textbf{TAMS}, however we  see that \textbf{TAMS-CLS} often outperforms \textbf{TAMS}. In fact, \textbf{TAMS-CLS} is often the highest performing model, which suggests that alignments are not strictly necessary to see performance gains from translation incorporation. We consider this a promising finding because of the relative difficulty of sourcing alignments.

\subsection{Manual vs. Automatic Alignment}

To directly compare the influence of automatic alignment on performance, we additionally train a \textbf{TAMS} model on gold-aligned Tsez data (\textbf{Gold}) and compare it to a \textbf{TAMS} model trained on equivalent data aligned with awesome-align. While we still report the average performance across 5 models trained on distinct training sets of 100 samples, the training sets we average here are distinct from the ones in \S \ref{subsec:test}) so performance metrics cannot be directly compared. To emphasize this point, we call the awesome-aligned setting \textbf{Awesome-Gold}. We report the performance of each model in Table \ref{tab:gold-test}. We find that \textbf{TAMS-CLS} is the highest performing model, suggesting that gold alignment data might not be necessary to see the best possible performance from \textbf{TAMS}. 

It is possible that the granularity of  information \textbf{TAMS} learns from alignments may in fact detract from its ability to generalize.  This would be a positive result because expert 
alignment is costly, and without the need for good alignments, training a \textbf{TAMS} model becomes
more feasible for a broader range of languages. However, further experimentation on more diverse languages is necessary to draw any conclusions to this end.

\begin{table}[t]
    \centering
    \small
    \begin{tabular}{c||c|cc} 
         \multicolumn{4}{c}{\textsc{Tsez}} \\ \midrule
         \textbf{Model} & \textbf{Metrics} & \textbf{Gold} & \textbf{Awesome-Gold} \\ \hline \hline
         & Accuracy & 32.68& 31.12\\ 
         TAMS & F1 & 48.50& 48.08\\ 
         & ED & \textbf{705}& 714\\
         \hline
 & Accuracy& \multicolumn{2}{c}{\textbf{33.96}}\\
 TAMS-CLS& F1& \multicolumn{2}{c}{\textbf{49.73}}\\
 & ED& \multicolumn{2}{c}{\textbf{705}}\\ \hline
         & Accuracy & \multicolumn{2}{c}{33.48} \\ 
         Baseline& F1 & \multicolumn{2}{c}{49.07} \\ 
         & ED & \multicolumn{2}{c}{727} \\ 
    \end{tabular}
    \caption{\textbf{Manual vs. automatic alignment:} Performance on Tsez's gold-aligned test split (n=100).}
    \label{tab:gold-test}
\end{table}

\subsection{Supplementary Analysis of Translation Incorporation Strategies}
\label{sec:translation incorporation}

In this section we provide details of the search that led to the final model configuration investigated above. We treat translation incorporation strategy as a hyperparameter and test all combinations on the Tsez development set to determine the optimal approach. 

We take the average whole-word accuracy for configuration across all training split sizes on the Tsez development set using the \textbf{CLS-None} strategy. The full results of this search are shown in Table \ref{table:mega-tsez}. From this search, we find that the overall highest performing strategy configuration is \textbf{Init-State} in the encoder and \textbf{Concat-Half} in the decoder. This configuration outperforms the baseline by an average of 1.58 percentage points and is the top performer on the 100 and 500 train sample settings. Only in the highest train data setting does the baseline outperform our top configuration. 

For the 100 train sample setting, \textbf{Concat} in the encoder and \textbf{Concat-Half} in the decoder, the second highest performing configuration overall, was the top configuration. With this in mind, we perform a second search over the CLS strategies on these two configurations, shown in Table \ref{table:Cls-tsez}.  From this, we find our final \textbf{TAMS} configuration: \textbf{Init-State} in the encoder, \textbf{Concat-Half} in the decoder and \textbf{CLS-Concat} as our CLS Strategy.

\begin{table*}[!ht]
    \centering
    \setlength{\tabcolsep}{4pt}
    \begin{tabular}{ll||lllll}
        \multicolumn{7}{c}{\textsc{Tsez}} \\ \midrule
        \textbf{\textbf{Encoder Strat.}} & \textbf{Decoder Strat.} & \textbf{n = 100}& \textbf{n = 250} & \textbf{n = 500} & \textbf{n = 6572} & \textbf{Average}\\ \hline \hline
        Init-State& Concat-Half & 22.47 & \textbf{34.02} & \textbf{45.39} & 80.46 & \textbf{45.58} \\ 
        Concat& Concat-Half & \textbf{24.66}& 32.33 & 45.02 & 80.09 & 45.53 \\ 
        Concat-Half& Concat-Half & 24.57 & 32.51 & 44.20 & 80.82 & 45.53 \\ 
        None& Concat-Half & 23.56 & 32.19 & 44.57 & 81.10 & 45.35 \\ 
        Concat& Init-State & 23.70 & 31.46 & 45.21 & 80.82 & 45.30 \\ 
        Concat& Concat & 22.92 & 32.05 & 44.84 & 80.55 & 45.09 \\ 
        Init-State& None & 22.60 & 31.46 & 44.43 & 80.91 & 44.85 \\ 
        None& Concat & 23.11 & 31.23 & 44.38 & 80.46 & 44.79 \\ 
        Concat& None & 23.24 & 30.18 & 45.21 & 80.46 & 44.77 \\ 
        Init-State& Init-State & 21.60 & 31.28 & 45.21 & 80.91 & 44.75 \\ 
        Concat-Half& None & 23.01 & 30.87 & 43.15 & 81.32 & 44.59 \\ 
        None& Init-State & 22.15 & 31.83 & 43.11 & 81.00 & 44.52 \\ 
        Concat-Half& Concat & 23.06 & 31.46 & 43.01 & 80.23 & 44.44 \\ 
        Init-State& Concat & 22.51 & 31.00 & 43.70 & 80.50 & 44.43 \\ 
        Concat-Half& Init-State & 22.88 & 31.42 & 40.96 & 81.32 & 44.14 \\ 
        \textit{None}& \textit{ None} & \textit{22.56}& \textit{28.90} & \textit{43.06} & \textit{\textbf{81.87}} & \textit{44.10} \\ 
        Init-Char& Concat-Half & 24.11 & 29.54 & 40.91 & 79.63 & 43.55 \\ 
        Init-Char& Concat & 21.74 & 31.23 & 39.54 & 79.27 & 42.95 \\ 
        Init-Char& Init-State & 22.33 & 24.89 & 41.55 & 80.09 & 42.21 \\ 
        Init-Char& None & 23.29 & 22.60 & 39.36 & 80.91 & 41.54 \\ 
    \end{tabular}
    \centering
    \caption{\textbf{Model tuning 1:} Accuracy (\%) of all translation incorporation strategies on Tsez's silver-aligned development split with no information from the CLS token (CLS-None).}
    \label{table:mega-tsez}
\end{table*}

\begin{table*}[!ht]
    \centering
    \setlength{\tabcolsep}{4pt}
    \begin{tabular}{lll||lllll}
        \multicolumn{8}{c}{\textsc{Tsez}} \\ \midrule
        \textbf{\textbf{Encoder Strat.}} & \textbf{\textbf{Decoder Strat.}} & \textbf{\textbf{CLS Strat.}} & \textbf{\textbf{n = 100}}& \textbf{\textbf{n = 250}}& \textbf{\textbf{n = 500}}& \textbf{\textbf{n = 6572}}& \textbf{\textbf{Average}} \\ \hline \hline
        Init-State & Concat-Half & CLS-Concat & 23.29 & 33.79 & 45.16 & 80.68 & \textbf{45.73}\\ 
        Init-State & Concat-Half & CLS-None & 22.47 & \textbf{34.02} & 45.39 & 80.46 & 45.58 \\ 
        Concat & Concat-Half & CLS-None & \textbf{24.66} & 32.33 & 45.02 & 80.09 & 45.53 \\ 
        Concat & Concat-Half & CLS-Avg & 23.70 & 32.42 & 45.30 & 80.41 & 45.46 \\ 
        Init-State & Concat-Half & CLS-Avg & 22.79 & 33.42 & 45.02 & 80.55 & 45.45 \\ 
        Concat & Concat-Half & CLS-Concat & 23.20 & 30.41 & \textbf{46.12} & 80.32 & 45.01  \\ 
        \textit{ None} & \textit{ None} & - & \textit{22.56}& \textit{28.90}& \textit{43.06}& \textit{\textbf{81.87}}& \textit{44.10} \\ 
    \end{tabular}
    \caption{\textbf{Model tuning 2:} Accuracy (\%) of CLS strategies with top-performing translation incorporation configurations on the Tsez silver-aligned development split.}
    \label{table:Cls-tsez}
\end{table*}

\section{Conclusion}
We present a novel method for incorporating information from translations into a morphological segmentation model to support low-resource canonical segmentation. 
Using Tsez as a development language, we determine our best-performing model (\textbf{TAMS}), which uses a fixed-length representation of the translation in two ways: to initialize the hidden state in the encoder (\textbf{Init-State}) and to concatenate to the input at each time step in the decoder (\textbf{Concat-Half}). 
Our model is most beneficial in the super low-resource setting (n=100), where it outperforms the baseline by  1.99 percentage points on average across three morphologically complex languages. 
And although we only tune our model on the Tsez development set, we also see performance gains for Arapaho, a typologically and morphologically distinct polysynthetic language. 
This promising result suggests that \textbf{TAMS} could be beneficial for a wide range of languages. We believe \textbf{TAMS} will be the most beneficial in the language documentation setting, where extreme resource constraints are realistic and often expected. 

The findings of our \textbf{TAMS-CLS} experiment are especially promising because they indicate that we may be able to see benefits from incorporating translations in canonical segmentation even without word-level alignments. This opens up TAMS to many more languages which may not have high-quality word-level alignments available.

However, our results are more mixed in higher-resource settings, which indicates that there is still more work to be done to determine whether translations are a valuable addition to canonical segmentation models in cases where more data is available. 
Overall, canonical segmentation for morphologically complex languages remains a challenging task, but we believe that this work indicates that translations should be explored further as an additional data resource.
There are several avenues for future work we wish to highlight.
A first possible improvement strategy could be to experiment with providing more explicit information instead of or in addition to translations, such as the POS tags of the aligned English words.
Second, it would be interesting to see whether our results can be reproduced on other languages and with other PLMs.
Third, there may be other ways to use translation-based representations with LSTMs.

\section*{Limitations} 
Due to data availability, we experimented only on two language families, Northeast Caucasian and Algonquian, but ideally we would have tested on more language families. We cannot concretely say that our models would perform equivalently on a more diverse set of languages. Another limitation was in the exhaustiveness of our hyperparameter search. Ideally, we would have searched each possible CLS token strategy with each possible configuration of translation incorporation strategies but we were unable to due to the GPU hours that would have been required.

\section{Acknowledgements}
We thank the anonymous reviewers for their useful suggestions and feedback, as well as the LECS and NALA Labs at the University of Colorado Boulder. We would like to extend our gratitude to Bernard Comrie and Sarah Moeller for their invaluable assistance with interpreting our datasets. Parts of this work were supported by the National Science
Foundation under Grant No. 2149404, “CAREER: From One Language to Another.” Any opinions, findings, and conclusions or recommendations expressed in this material are those of the authors and do not necessarily reflect the views of the National
Science Foundation. This work utilized the Blanca condo computing resource at the University of Colorado Boulder. Blanca is jointly funded by computing users and the University of Colorado Boulder. 

% Entries for the entire Anthology, followed by custom entries
\bibliography{anthology,custom}

\begin{thebibliography}{30}
\expandafter\ifx\csname natexlab\endcsname\relax\def\natexlab#1{#1}\fi

\bibitem[{Ahumada et~al.(2022)Ahumada, Gutierrez, and Anastasopoulos}]{ahumada-etal-2022-educational}
Cristian Ahumada, Claudio Gutierrez, and Antonios Anastasopoulos. 2022.
\newblock \href {https://doi.org/10.18653/v1/2022.bea-1.23} {Educational tools for mapuzugun}.
\newblock In \emph{Proceedings of the 17th Workshop on Innovative Use of NLP for Building Educational Applications (BEA 2022)}, pages 183--196, Seattle, Washington. Association for Computational Linguistics.

\bibitem[{Bahdanau et~al.(2015)Bahdanau, Cho, and Bengio}]{bahdanau2015neural}
Dzmitry Bahdanau, Kyunghyun Cho, and Yoshua Bengio. 2015.
\newblock Neural machine translation by jointly learning to align and translate.
\newblock In \emph{ICLR}.

\bibitem[{Chaudhary et~al.(2022)Chaudhary, Sheikh, Mortensen, Anastasopoulos, and Neubig}]{chaudhary2022autolex}
Aditi Chaudhary, Zaid Sheikh, David~R Mortensen, Antonios Anastasopoulos, and Graham Neubig. 2022.
\newblock \href {http://arxiv.org/abs/2203.13901} {Autolex: An automatic framework for linguistic exploration}.

\bibitem[{Comrie and Polinsky()}]{comrie_and_polinksy}
Bernard Comrie and Maria Polinsky.
\newblock Tsez.
\newblock Unpublished manuscript.

\bibitem[{Devlin et~al.(2019)Devlin, Chang, Lee, and Toutanova}]{Devlin_Chang_Lee_Toutanova_2019}
Jacob Devlin, Ming-Wei Chang, Kenton Lee, and Kristina Toutanova. 2019.
\newblock \href {https://doi.org/10.18653/v1/N19-1423} {Bert: Pre-training of deep bidirectional transformers for language understanding}.
\newblock In \emph{Proceedings of the 2019 Conference of the North American Chapter of the Association for Computational Linguistics: Human Language Technologies, Volume 1 (Long and Short Papers)}, page 4171–4186, Minneapolis, Minnesota. Association for Computational Linguistics.

\bibitem[{Dou and Neubig(2021)}]{Dou_Neubig_2021}
Zi-Yi Dou and Graham Neubig. 2021.
\newblock \href {https://doi.org/10.18653/v1/2021.eacl-main.181} {Word alignment by fine-tuning embeddings on parallel corpora}.
\newblock In \emph{Proceedings of the 16th Conference of the European Chapter of the Association for Computational Linguistics: Main Volume}, page 2112–2128, Online. Association for Computational Linguistics.

\bibitem[{Ginn et~al.(2023)Ginn, Moeller, Palmer, Stacey, Nicolai, Hulden, and Silfverberg}]{Ginn_Moeller_Palmer_Stacey_Nicolai_Hulden_Silfverberg_2023}
Michael Ginn, Sarah Moeller, Alexis Palmer, Anna Stacey, Garrett Nicolai, Mans Hulden, and Miikka Silfverberg. 2023.
\newblock \href {https://doi.org/10.18653/v1/2023.sigmorphon-1.20} {Findings of the sigmorphon 2023 shared task on interlinear glossing}.
\newblock In \emph{Proceedings of the 20th SIGMORPHON workshop on Computational Research in Phonetics, Phonology, and Morphology}, page 186–201, Toronto, Canada. Association for Computational Linguistics.

\bibitem[{Haspelmath(1993)}]{Haspelmath_1993}
Martin Haspelmath. 1993.
\newblock \emph{A grammar of Lezgian}.
\newblock Mouton de Gruyter.

\bibitem[{Jawahar et~al.(2019)Jawahar, Sagot, and Seddah}]{jawahar-etal-2019-bert}
Ganesh Jawahar, Beno{\^\i}t Sagot, and Djam{\'e} Seddah. 2019.
\newblock \href {https://doi.org/10.18653/v1/P19-1356} {What does {BERT} learn about the structure of language?}
\newblock In \emph{Proceedings of the 57th Annual Meeting of the Association for Computational Linguistics}, pages 3651--3657, Florence, Italy. Association for Computational Linguistics.

\bibitem[{Kann et~al.(2016)Kann, Cotterell, and Schütze}]{Kann_Cotterell_Schütze_2016}
Katharina Kann, Ryan Cotterell, and Hinrich Schütze. 2016.
\newblock \href {https://doi.org/10.18653/v1/D16-1097} {Neural morphological analysis: Encoding-decoding canonical segments}.
\newblock In \emph{Proceedings of the 2016 Conference on Empirical Methods in Natural Language Processing}, page 961–967, Austin, Texas. Association for Computational Linguistics.

\bibitem[{Lehmann(1982)}]{Lehmann_1982}
Christian Lehmann. 1982.
\newblock \href {https://doi.org/10.1515/flin.1982.16.1-4.199} {Directions for interlinear morphemic translations}.
\newblock 16(1–4):199–224.

\bibitem[{Mager et~al.(2020)Mager, Çetinoğlu, and Kann}]{Mager_Çetinoğlu_Kann_2020}
Manuel Mager, Özlem Çetinoğlu, and Katharina Kann. 2020.
\newblock \href {https://doi.org/10.18653/v1/2020.emnlp-main.423} {Tackling the low-resource challenge for canonical segmentation}.
\newblock In \emph{Proceedings of the 2020 Conference on Empirical Methods in Natural Language Processing (EMNLP)}, page 5237–5250, Online. Association for Computational Linguistics.

\bibitem[{Melamed(1998)}]{Melamed_1998}
I.~Dan Melamed. 1998.
\newblock \href {http://arxiv.org/abs/cmp-lg/9805004} {Annotation style guide for the blinker project}.
\newblock (arXiv:cmp-lg/9805004).
\newblock ArXiv:cmp-lg/9805004.

\bibitem[{Moeller and Hulden(2021)}]{moeller-hulden-2021-integrating}
Sarah Moeller and Mans Hulden. 2021.
\newblock \href {https://aclanthology.org/2021.computel-1.11} {Integrating automated segmentation and glossing into documentary and descriptive linguistics}.
\newblock In \emph{Proceedings of the 4th Workshop on the Use of Computational Methods in the Study of Endangered Languages Volume 1 (Papers)}, pages 86--95, Online. Association for Computational Linguistics.

\bibitem[{Moeller et~al.(2020)Moeller, Liu, Yang, Kann, and Hulden}]{moeller-etal-2020-igt2p}
Sarah Moeller, Ling Liu, Changbing Yang, Katharina Kann, and Mans Hulden. 2020.
\newblock \href {https://doi.org/10.18653/v1/2020.emnlp-main.424} {{IGT}2{P}: From interlinear glossed texts to paradigms}.
\newblock In \emph{Proceedings of the 2020 Conference on Empirical Methods in Natural Language Processing (EMNLP)}, pages 5251--5262, Online. Association for Computational Linguistics.

\bibitem[{Moeng et~al.(2021)Moeng, Reay, Daniels, and Buys}]{Moeng_Reay_Daniels_Buys_2021}
Tumi Moeng, Sheldon Reay, Aaron Daniels, and Jan Buys. 2021.
\newblock \href {https://ui.adsabs.harvard.edu/abs/2021arXiv210400767M type: article} {\emph{Canonical and Surface Morphological Segmentation for Nguni Languages}}.

\bibitem[{Mohseni and Tebbifakhr(2019)}]{Mohseni_Tebbifakhr_2019}
Mahdi Mohseni and Amirhossein Tebbifakhr. 2019.
\newblock \href {https://aclanthology.org/2019.nsurl-1.4} {Morphobert: a persian ner system with bert and morphological analysis}.
\newblock In \emph{Proceedings of the First International Workshop on NLP Solutions for Under Resourced Languages (NSURL 2019) co-located with ICNLSP 2019 - Short Papers}, page 23–30, Trento, Italy. Association for Computational Linguistics.

\bibitem[{Musil(2021)}]{musil-2021-representations}
Tom{\'a}{\v{s}} Musil. 2021.
\newblock \href {https://doi.org/10.18653/v1/2021.naacl-srw.4} {Representations of meaning in neural networks for {NLP}: a thesis proposal}.
\newblock In \emph{Proceedings of the 2021 Conference of the North American Chapter of the Association for Computational Linguistics: Student Research Workshop}, pages 24--31, Online. Association for Computational Linguistics.

\bibitem[{Nastase and Merlo(2023)}]{nastase-merlo-2023-grammatical}
Vivi Nastase and Paola Merlo. 2023.
\newblock \href {https://doi.org/10.18653/v1/2023.repl4nlp-1.3} {Grammatical information in {BERT} sentence embeddings as two-dimensional arrays}.
\newblock In \emph{Proceedings of the 8th Workshop on Representation Learning for NLP (RepL4NLP 2023)}, pages 22--39, Toronto, Canada. Association for Computational Linguistics.

\bibitem[{Palmer et~al.(2009)Palmer, Moon, and Baldridge}]{Palmer_Moon_Baldridge_2009}
Alexis Palmer, Taesun Moon, and Jason Baldridge. 2009.
\newblock \href {https://aclanthology.org/W09-1905} {Evaluating automation strategies in language documentation}.
\newblock In \emph{Proceedings of the NAACL HLT 2009 Workshop on Active Learning for Natural Language Processing}, page 36–44, Boulder, Colorado. Association for Computational Linguistics.

\bibitem[{Ruzsics and Samard{\v{z}}i{\'c}(2017)}]{ruzsics-samardzic-2017-neural}
Tatyana Ruzsics and Tanja Samard{\v{z}}i{\'c}. 2017.
\newblock \href {https://doi.org/10.18653/v1/K17-1020} {Neural sequence-to-sequence learning of internal word structure}.
\newblock In \emph{Proceedings of the 21st Conference on Computational Natural Language Learning ({C}o{NLL} 2017)}, pages 184--194, Vancouver, Canada. Association for Computational Linguistics.

\bibitem[{Schone and Jurafsky(2001)}]{schone-jurafsky-2001-knowledge}
Patrick Schone and Daniel Jurafsky. 2001.
\newblock \href {https://aclanthology.org/N01-1024} {Knowledge-free induction of inflectional morphologies}.
\newblock In \emph{Second Meeting of the North {A}merican Chapter of the Association for Computational Linguistics}.

\bibitem[{See et~al.(2017)See, Liu, and Manning}]{see-etal-2017-get}
Abigail See, Peter~J. Liu, and Christopher~D. Manning. 2017.
\newblock \href {https://doi.org/10.18653/v1/P17-1099} {Get to the point: Summarization with pointer-generator networks}.
\newblock In \emph{Proceedings of the 55th Annual Meeting of the Association for Computational Linguistics (Volume 1: Long Papers)}, pages 1073--1083, Vancouver, Canada. Association for Computational Linguistics.

\bibitem[{Singh et~al.(2021)Singh, Rutten, and Lefever}]{singh-etal-2021-pilot}
Pranaydeep Singh, Gorik Rutten, and Els Lefever. 2021.
\newblock \href {https://doi.org/10.18653/v1/2021.latechclfl-1.15} {A pilot study for {BERT} language modelling and morphological analysis for ancient and medieval {G}reek}.
\newblock In \emph{Proceedings of the 5th Joint SIGHUM Workshop on Computational Linguistics for Cultural Heritage, Social Sciences, Humanities and Literature}, pages 128--137, Punta Cana, Dominican Republic (online). Association for Computational Linguistics.

\bibitem[{Soricut and Och(2015)}]{soricut-och-2015-unsupervised}
Radu Soricut and Franz Och. 2015.
\newblock \href {https://doi.org/10.3115/v1/N15-1186} {Unsupervised morphology induction using word embeddings}.
\newblock In \emph{Proceedings of the 2015 Conference of the North {A}merican Chapter of the Association for Computational Linguistics: Human Language Technologies}, pages 1627--1637, Denver, Colorado. Association for Computational Linguistics.

\bibitem[{Tsai et~al.(2019)Tsai, Riesa, Johnson, Arivazhagan, Li, and Archer}]{tsai-etal-2019-small}
Henry Tsai, Jason Riesa, Melvin Johnson, Naveen Arivazhagan, Xin Li, and Amelia Archer. 2019.
\newblock \href {https://doi.org/10.18653/v1/D19-1374} {Small and practical {BERT} models for sequence labeling}.
\newblock In \emph{Proceedings of the 2019 Conference on Empirical Methods in Natural Language Processing and the 9th International Joint Conference on Natural Language Processing (EMNLP-IJCNLP)}, pages 3632--3636, Hong Kong, China. Association for Computational Linguistics.

\bibitem[{Vinyals et~al.(2015)Vinyals, Fortunato, and Jaitly}]{Vinyals_Fortunato_Jaitly_2015}
Oriol Vinyals, Meire Fortunato, and Navdeep Jaitly. 2015.
\newblock \href {https://proceedings.neurips.cc/paper_files/paper/2015/file/29921001f2f04bd3baee84a12e98098f-Paper.pdf} {Pointer networks}.
\newblock In \emph{Advances in Neural Information Processing Systems}, volume~28. Curran Associates, Inc.

\bibitem[{Wolf et~al.(2020)Wolf, Debut, Sanh, Chaumond, Delangue, Moi, Cistac, Rault, Louf, Funtowicz, Davison, Shleifer, von Platen, Ma, Jernite, Plu, Xu, Le~Scao, Gugger, Drame, Lhoest, and Rush}]{Wolf_Debut_Sanh_Chaumond_Delangue_Moi_Cistac_Rault_Louf_Funtowicz_et_al._2020}
Thomas Wolf, Lysandre Debut, Victor Sanh, Julien Chaumond, Clement Delangue, Anthony Moi, Pierric Cistac, Tim Rault, Remi Louf, Morgan Funtowicz, Joe Davison, Sam Shleifer, Patrick von Platen, Clara Ma, Yacine Jernite, Julien Plu, Canwen Xu, Teven Le~Scao, Sylvain Gugger, Mariama Drame, Quentin Lhoest, and Alexander Rush. 2020.
\newblock \href {https://doi.org/10.18653/v1/2020.emnlp-demos.6} {Transformers: State-of-the-art natural language processing}.
\newblock In \emph{Proceedings of the 2020 Conference on Empirical Methods in Natural Language Processing: System Demonstrations}, page 38–45, Online. Association for Computational Linguistics.

\bibitem[{Yarowsky and Wicentowski(2000)}]{yarowsky-wicentowski-2000-minimally}
David Yarowsky and Richard Wicentowski. 2000.
\newblock \href {https://doi.org/10.3115/1075218.1075245} {Minimally supervised morphological analysis by multimodal alignment}.
\newblock In \emph{Proceedings of the 38th Annual Meeting of the Association for Computational Linguistics}, pages 207--216, Hong Kong. Association for Computational Linguistics.

\bibitem[{Zhao et~al.(2020)Zhao, Ozaki, Anastasopoulos, Neubig, and Levin}]{Zhao_Ozaki_Anastasopoulos_Neubig_Levin_2020}
Xingyuan Zhao, Satoru Ozaki, Antonios Anastasopoulos, Graham Neubig, and Lori Levin. 2020.
\newblock \href {https://doi.org/10.18653/v1/2020.coling-main.471} {Automatic interlinear glossing for under-resourced languages leveraging translations}.
\newblock In \emph{Proceedings of the 28th International Conference on Computational Linguistics}, page 5397–5408, Barcelona, Spain (Online). International Committee on Computational Linguistics.

\end{thebibliography}
\bibliographystyle{acl_natbib}

\onecolumn
\appendix

\section{Manual Word Alignment Directives}
\label{sec:appendixa}
\subsection{High-level Directives}
\begin{itemize}
    \item Align English prepositions with the Tsez word with the equivalent case marker.
    \item If some grammatical information is expressed in one language but not the other, behave as if it were expressed in the corresponding phrase. In such cases, the extra word (`the', for example) should be aligned to the head of the corresponding phrase.
    \item But, if the definiteness is expressed on a modifier of the noun head (like an attributive adjective), then align the article to the modifier bearing the definiteness information instead.
\end{itemize}
\subsection{Lower-level Directives}
\begin{itemize}
    \item Pronominal subjects that are not expressed in Tsez and are expressed in English should have the English subject aligned to the head predicate in Tsez.
    \item If in English you have a PP and in Tsez you have a relative clause where the subject position has the gap, align the preposition introducing the PP with the relative clause’s predicate
    \item If a quotative verb like `say' is used a variable amount of times in one language compared to the other, then align all instances of the quotative verbs together, so long as the quoted material they’re all referring to is identical.
    \item Always align the expletive `there' with the existential verb in Tsez. And if there is an adverbial `there'-equivalent in Tsez, do not align it with anything in English unless there really is an adverbial, non-expletive `there' or similar in English
    \item For articles in English, if there’s something very close to an article in Tsez (‘a’, or ‘this’, or …), then prefer aligning the English article to the similar word instead of the noun.

\end{itemize}

\section{Hyperparameter Tuning}
\label{sec:appendixb}
We conduct hyperparameter tuning for the baseline Tsez models without translation using random search with the full training set of 6572  words. We then took the top two architectures and performed a hyper-parameter sweep with 100 and 500 training samples to simulate a lower-resource setting. The top performing models for each architecture and training set size are outlined in Tables \ref{fig:100}, \ref{fig:500}, \ref{fig:full}. The search space of architecture specific hyperparameters is outlined in  Table \ref{fig:arch} and Table \ref{fig:cond} and the search space of optimization parameters is outlined in Table \ref{fig:opt}. All models used Adam optimization. We report whole-word accuracy on the development set.                                                         
\begin{table}[h]
    \centering
    \caption{Best Performing Hyperparameters for Each Architecture with 6572 Training Samples}
    \small
    \begin{tabular}{|l|c|c|c|}
        \hline
        \textbf{Hyperparameter} & \textbf{Transformer} & \textbf{Pointer-Generator LSTM} & \textbf{Attentive LSTM} \\
        \hline
        Embedding Size & 512& 896& 512\\
        Hidden Size & 1024 & 1856& 960\\
        Dropout & 0.3022& 0.2212& 0.07615\\
        Attention Heads & 8 & 1& 1 \\
        Encoder Layers & 4& 1& 2 \\
 Decoder Layers& 2& 1&1\\
        Batch Size & 16& 64 & 128\\
        Learning Rate (LR)& 0.0001975& 0.0008056& 0.0002227\\
        Beta1 & 0.8153& 0.8218& 0.841\\
        Beta2 & 0.9874& 0.9845& 0.9815\\
        Scheduler & reduceonplateau& reduceonplateau&  None \\
        Num Warmup Samples & -& -& -\\
        Reduce LR Factor& 0.3095& 0.782& -\\
        Reduce LR Patience & 40& 30& -\\
        Min LR &  0.3095& 0.0007737& -\\
        \hline
        \hline
        \textbf{Accuracy} & 0.8629 & 0.8634 & 0.8645 \\
        \hline 
    \end{tabular}
    \label{fig:full}
\end{table}

\begin{table}[h]
    \centering
    \caption{Best Performing Hyperparameters for Each Architecture with 500 Training Samples}
    \small
    \begin{tabular}{|l|c|c|}
        \hline
        \textbf{Hyperparameter} & \textbf{Pointer-Generator LSTM} & \textbf{Attentive LSTM} \\
        \hline
        Embedding Size & 320& 320\\
        Hidden Size & 1728& 2048\\
        Dropout & 0.3915& 0.4794\\
        Attention Heads & 1& 1\\
        Encoder Layers & 1& 2\\
        Decoder Layers & 1& 1\\
        Batch Size & 64& 16\\
        Learning Rate & 0.0007847& 0.00008051\\
        Beta1 & 0.8699& 0.8789\\
        Beta2 & 0.9803& 0.9971\\
        Scheduler & -& -\\
        Num Warmup Samples & -& -\\
        Reduce LR Factor& -& -\\
        Reduce LR Patience & -& -\\
        Min LR & -& -\\
        \hline
        \hline
        \textbf{Accuracy} & 0.5059& 0.5059\\
        \hline 
    \end{tabular}
    \label{fig:500}
\end{table}

\begin{table}[h]
    \centering
    \caption{Best Performing Hyperparameters for Each Architecture with 100 Training Samples}
    \small
    \begin{tabular}{|l|c|c|}
        \hline
        \textbf{Hyperparameter} & \textbf{Pointer-Generator LSTM} & \textbf{Attentive LSTM} \\
        \hline
        Embedding Size & 640& 192\\
        Hidden Size & 896& 384\\
        Dropout & 0.3662& 0.3132\\
        Attention Heads & 1& 1\\
        Encoder Layers & 2& 1\\
        Decoder Layers & 2& 1\\
        Batch Size & 16& 16\\
        Learning Rate & 0.0002411& 0.0000523\\
        Beta1 & 0.8716& 0.8263\\
        Beta2 & 0.9848& 0.9875\\
        Scheduler & 'reduceonplateau' & -\\
        Num Warmup Samples & -& -\\
        Reduce LR Factor& 0.686& -\\
        Reduce LR Patience & 30& -\\
        Min LR & 0.0005021& -\\
        \hline
        \hline
        \textbf{Accuracy} & 0.2409& 0.157\\
        \hline 
    \end{tabular}
    \label{fig:100}
\end{table}

\begin{table}[h]
    \centering
    \caption{Architecture Hyperparameters Search Space}
    \begin{tabular}{|l|c|c|c|}
        \hline
        \textbf{Hyperparameter} & \textbf{Distribution} & \textbf{Minimum} & \textbf{Maximum} \\
        \hline
        Embedding Size & q\_uniform & 128 & 1024 \\
        Hidden Size & q\_uniform & 128 & 2048 \\
        Dropout & uniform & 0 & 0.5 \\
        \hline
    \end{tabular}
    \label{fig:arch}
\end{table}

\begin{table}[h]
    \centering
    \caption{Conditional Hyperparameters based on Architecture Type}
    \begin{tabular}{|l|c|c|c|c|}
        \hline
        \textbf{Model} & \textbf{Attention Heads} & \textbf{Number of Encoder Layers} &  \textbf{Number of Decoder Layers}\\
        \hline
        Transformer & [2, 4, 8] & [2, 4, 6, 8] & [2, 4, 6, 8] \\
        Pointer-Generator LSTM & [1] & [1, 2] & [1, 2]\\
        Attentive LSTM & [1] & [1, 2]& [1, 2]\\
        \hline
    \end{tabular}
    \label{fig:cond}
\end{table}

\begin{table}[h]
    \centering
    \caption{Optimization Hyperparameters Search Space}
    \begin{tabular}{|l|c|c|}
        \hline
        \textbf{Hyperparameter} & \textbf{Distribution} & \textbf{Values} \\
        \hline
        Batch Size & categorical & [16, 32, 64] \\
        Learning Rate & log\_uniform\_values & $1 \times 10^{-6}$ to 0.01 \\
        Beta1 & uniform & 0.8 to 0.999 \\
        Beta2 & uniform & 0.98 to 0.999 \\
        Scheduler & values & ['reduceonplateau', 'warmupinvsqrt',  None] \\
        Num Warmup Samples & q\_uniform & 0 to 5000000 \\
        Reduce LR Factor & uniform & 0.1 to 0.9 \\
        Reduce LR Patience & q\_uniform & 10 to 50 \\
        Min LR & uniform & $1 \times 10^{-7}$ to 0.001 \\
        \hline
    \end{tabular}
    \label{fig:opt}
\end{table}

\label{appendixc}
\section{Standard Deviation of TAMS Performance}

\begin{table*}[t]
    \centering
    \scriptsize
    \begin{tabular}{cc|ccc|ccc|ccc|ccc} 
    & & \multicolumn{3}{c}{\textsc{Tsez}} & \multicolumn{3}{c}{\textsc{Lezgi}} & \multicolumn{3}{c}{\textsc{Arapaho}} & \multicolumn{3}{c}{\textsc{Average}}  \\ \midrule
    \textbf{Train} & \textbf{Metrics} & \textbf{TAMS} & \textbf{TAMS-} & \textbf{Base} & \textbf{TAMS} & \textbf{TAMS-} & \textbf{Base} & \textbf{TAMS} & \textbf{TAMS-} & \textbf{Base} & \textbf{TAMS} & \textbf{TAMS-} & \textbf{Base} \\
    \textbf{Limit} & & & \textbf{CLS} & & & \textbf{CLS} & & & \textbf{CLS} & & & \textbf{CLS} \\
    \midrule
    & Acc. & 2.52 & 2.92 & 2.18 & 2.27 & 0.99 & 1.30 & 2.21 & 2.87 & 2.05 & 2.33 & 2.26 & 1.84 \\
    n = & F1 & 2.87 & 2.83 & 2.23 & 3.57 & 3.70 & 4.10 & 3.04 & 3.73 & 3.43 & 3.16 & 3.42 & 3.25 \\ 
    100 & ED & 155 & 221 & 101 & 33 & 20 & 37 & 3006 & 3334 & 3727 & 1065 & 1192 & 1288 \\  \hline
    & Acc. &  1.62 & 1.77 & 0.93 & 1.96 & 2.07 & 1.51 & 1.30 & 1.75 & 0.96 & 1.63 & 1.86 & 1.11 \\  
    n = & F1 &  1.53 & 1.62 & 0.94 & 1.78 & 1.54 & 2.85 & 1.40 & 1.86 & 0.90 & 1.57 & 2.00 & 1.57 \\ 
    250 & ED & 117 & 152 & 165 & 23 & 31 & 29 & 1961 & 1292 & 1536 & 700 & 492 & 577 \\ \hline
    & Acc. & 1.54 & 1.78 & 0.89 & 2.37 & 1.92 & 2.15 & 1.59 & 1.69 & 1.28 & 1.83 & 1.80 & 1.44 \\  
    n = & F1 & 1.07 & 1.09 & 0.67 & 1.85 & 1.48 & 1.57 & 1.32 & 1.08 & 0.67 & 1.41 & 1.21 & 0.97 \\ 
    500 & ED & 110 & 120 & 64 & 27 & 24 & 31 & 688 & 967 & 918 & 275 & 370 & 338 \\ 
    \bottomrule
    \end{tabular}
    \caption{\textbf{Standard deviation results per language:} Standard deviation of performance metrics on all languages' test sets (over 5 randomized test sets) on silver-aligned data. Metrics: Accuracy (Acc.), F1 score (F1), and Edit Distance (ED).}
    \label{tab:combined-test-std}
\end{table*}

\begin{table*}[t]
    \centering
    \small
    \begin{tabular}{c||c|cc} 
         \multicolumn{4}{c}{\textsc{Tsez}} \\ \midrule
         \textbf{Model} & \textbf{Metrics} & \textbf{Gold} & \textbf{Awesome-Gold} \\ \hline \hline
         & Accuracy & 1.40& 2.48\\ 
         TAMS & F1 & 2.16& 2.14\\ 
         & ED & 33& 42\\
         \hline
 & Accuracy& \multicolumn{2}{c}{1.48}\\
 TAMS-CLS& F1& \multicolumn{2}{c}{2.14}\\
 & ED& \multicolumn{2}{c}{33}\\ \hline
         & Accuracy & \multicolumn{2}{c}{1.98} \\ 
         Baseline& F1 & \multicolumn{2}{c}{2.56} \\ 
         & ED & \multicolumn{2}{c}{48} \\ 
    \end{tabular}
    \caption{Standard deviation results on Tsez's gold-aligned test split (n=100).}
    \label{tab:gold-std}
\end{table*}

\end{document}